# Assessing LLMs' Performance: Insights from the Chinese Pharmacist Exam

Xinran WANG[1], Boran ZHU[2], Shujuan ZHOU[3], Ziwen LONG[4]

Dehua ZHOU[†], Shu ZHANG[†]

[1]Master of Public Health, City University of Hong Kong, Hong Kong SAR, China
[2]School of Chinese Medicine, Nanjing University of Chinese Medicine, Nanjing, China
[3]Department of Radiation Oncology, Fudan University Shanghai Cancer Center, Shanghai, China
[4]Department of Gastric Cancer, Fudan University Shanghai Cancer Center, Shanghai, China

[†]Corresponding authors

Dehua ZHOU (deachzdh@163.com), Department of Gastrointestinal Surgery, Shanghai Fourth People's Hospital Affiliated to Tongji University School of Medicine, Shanghai 200434, China

Shu ZHANG (superdrzhang@yeah.net), Department of Oncology, Fudan University Shanghai Cancer Center, Shanghai, China

**Abstract**

**Background**

As large language models (LLMs) become increasingly integrated into digital health education and assessment workflows, their capabilities in supporting high-stakes, domain-specific certification tasks remain underexplored.

In China, the national pharmacist licensure exam serves as a standardized benchmark for evaluating pharmacists' clinical and theoretical competencies.

**Objective**

This study aimed to compare the performance of two LLMs—ChatGPT-4o and DeepSeek-R1—on real questions from the Chinese Pharmacist Licensing Examination (2017 – 2021), and to discuss the implications of these performance differences for AI-enabled formative evaluation.

**Methods**

A total of 2,306 multiple-choice questions (text-only) questions were compiled from official exams, training materials, and public databases. Questions containing tables or images were excluded. Each item was input in its original Chinese format, and model responses were evaluated for exact accuracy. Pearson's Chi-squared test was used to compare overall performance, and Fisher's exact test was applied to year-wise multiple-choice accuracy.

**Results**

DeepSeek-R1 outperformed ChatGPT-4o with a significantly higher overall accuracy (90.0 % vs. 76.1 %, $p < 0.001$). Unit-level analyses revealed consistent advantages for DeepSeek-R1, particularly in foundational and clinical synthesis modules. While year-by-year multiple-choice performance also favored DeepSeek-R1, this performance gap did not reach statistical significance in any specific unit-year (all $p > .05$).

**Conclusion**

DeepSeek-R1 demonstrated robust alignment with the structural and semantic demands of the pharmacist licensure exam. These findings suggest that domain-specific models warrant further investigation for this context, while also reinforcing the necessity of human oversight in legally and ethically sensitive contexts.

**Keywords**

Large language models; Pharmacist licensure examination; ChatGPT-4o; DeepSeek-R1; Pharmacy education; Intelligent assessment; Healthcare informatics

**Introduction**

With the growing demand for healthcare services in China, the need for qualified pharmacists has increased significantly. In response, since 2008, 25 universities have been authorized to offer Bachelor's, Master's, and Ph.D. programs in clinical pharmacy, with durations ranging from 3 to 7 years[1]. In China, clinical pharmacy is an essential component of the public healthcare system, closely connected with its subsystems of education, training, qualification authentication, scientific research, management, and human resources[2]. Chinese clinical pharmacists provide patient-centered services including medication education, prescription review, and adverse drug reaction (ADR) monitoring. They also offer medication and laboratory recommendations, and in some cases, provide consultation to other healthcare professionals[3]. During the COVID-19 pandemic, pharmacists in China not only provided essential pharmaceutical care to the general public[4], but also provided assistance to cabin hospitals via remote internet platforms. Their responsibilities further included drug therapy monitoring, consultation and education to prevent and identify drug-related problems[5], as well as public health education services[6].

As the role of pharmacists continues to evolve, the ability to make sound professional decisions has become essential. In this context, China's pharmacist licensure exam plays a central role. It isn't just a hurdle to clear—it reflects the kinds of real-world decisions pharmacists are expected to make, whether dealing with patients, handling medications, or working within legal and institutional frameworks. The exam acts as a gatekeeper, setting a clear baseline to ensure that those entering the profession are prepared for the practical and ethical demands they will face in everyday work.

The 7th edition of China's pharmacist licensure examination syllabus places a strong emphasis on improving the overall quality of the pharmacist workforce. It aims to promote the healthy and standardized development of the profession, not only by testing core knowledge, but also by aligning assessment content with evolving healthcare needs and responsibilities[7].

In parallel with these developments in pharmacy practice, a significant technological shift is occurring. Large language models (LLMs) have been attracting attention across many fields[8], from architecture[9] to education[10], and from general medicine[11] to healthcare services[12]. Tools like ChatGPT-4o, in particular, have shown that they can handle a wide range of tasks—answering questions[13], explaining reasoning[14], even simulating medical exams in specialized areas like oncology[15].

This intersection of a large, evolving pharmacist workforce in China and the rise of powerful AI models creates a critical area of inquiry. The national licensure exam sits at the center of this context: it not only validates the workforce but also serves as a robust benchmark to test the domain-specific capabilities of LLMs.

In this study, we wanted to see how two well-known language models — ChatGPT-4o, a general-purpose model[16], and DeepSeek-R1, a Chinese open-source model[17] trained with domain-specific materials—would perform when faced with the real questions used in China's national pharmacist exam.

Our study builds upon recent work where LLMs such as ChatGPT-4o have been evaluated on various professional and academic assessments[18], including the United States Medical Licensing Examination (USMLE)[19], the bar exam[20], and college entrance tests[21]. While results vary by subject and format, these studies highlight LLMs' emerging ability to process domain-specific language and simulate human-like test-taking behavior[22] — raising the possibility of their integration into education and assessment workflows.

However, the current literature suffers from a significant limitation: it is overwhelmingly concentrated on the English language and Western assessment frameworks (such as the USMLE). This English-centric bias severely limits our understanding of how LLMs perform in different linguistic, cultural, and regulatory contexts, especially within non-English, high-stakes certification systems.

Therefore, this study does not merely extend existing benchmarks; it challenges them by asking a critical question: Can the strong performance of a general-purpose model like ChatGPT-4o be replicated in an environment that is linguistically and professionally distinct?

Furthermore, by directly comparing a general-purpose model (ChatGPT-4o) against a model specifically optimized for the domain and language (DeepSeek-R1), this study investigates the extent to which "general intelligence" translates to "specialized professional competence." Evaluating their performance across different exam units not only reveals their technical capabilities but also provides the empirical foundation necessary to discuss the unique pedagogical risks, fairness considerations, and interpretability challenges associated with deploying AI in a Chinese educational context.

## Methods

### Data Source and Preparation

This study examined how two different methods performed when applied to a set of questions from the Chinese National Pharmacist Licensing Examinations (NPLE), covering the years 2017 to 2021. A total of 2,306 questions were included.

We selected 2017 – 2021 exam items to reduce the chance of overlap with LLM training data. Questions from later years are more likely to have circulated online and thus potentially be seen by models like ChatGPT-4o or DeepSeek-R1.

These were collected from official exam notices, commonly used preparation books, and publicly available question banks. Before the analysis began, the dataset was carefully cleaned. Duplicate and incomplete entries were removed, and each question was reviewed by hand to check for clarity and consistent formatting. Questions that involved tables or images were left out, in order to keep the input straightforward and focused on textual content.

### Question Format and Classification

The NPLE dataset covers four main professional units, each testing a different aspect of pharmaceutical competence. These units are: Unit 1 (Pharmaceutical Foundations), which tests core factual knowledge in areas like pharmacology, chemistry, and pharmaceutics, often requiring straightforward recall; Unit 2 (Pharmaceutical Legislation and Ethics), which focuses on the interpretation of legal language, professional ethics, and region-specific healthcare regulations; Unit 3 (Dispensing and Prescription Review), which assesses practical, case-based judgment in real-world scenarios, such as identifying drug interactions or patient counseling needs; and Unit 4 (Clinical Integration), which involves comprehensive, case-based learning, requiring candidates to synthesize multiple pieces of information to plan and evaluate patient treatment paths. Our total dataset comprised 2,306 questions from these units (2,114 single-choice and 192 multiple-choice items). Questions were classified by both their unit and their format (single-choice vs. multiple-choice) to allow for a granular analysis of model performance.

**Model Input and Answer Evaluation**

To ensure methodological transparency, we detail our implementation strategy. A single, consistent system-level prompt was provided to both models to set the task context. The exact prompt used was:

" *You are a medical expert tasked with answering official licensing exam questions. Read each question carefully and provide the most accurate and evidence-based response. If the question requires clinical reasoning, follow established medical guidelines. If the question involves pharmacy, use pharmacological principles and regulatory knowledge.*"

Immediately following this prompt, each question (consisting of the question stem and all answer options) was input as a single, concatenated text block.

Each question was presented to the two models in its original Chinese format. To maintain uniformity, models were not provided with additional context beyond the question stem and answer options. Model outputs were matched against the official answer keys. A response was marked correct only if it matched the standard answer without ambiguity. For multiple-choice questions, full accuracy was required—partial correctness was not considered acceptable.

**Statistical Analysis**

All statistical analyses were performed using R (Version 4.3.2). A significance threshold was set at $p < 0.05$. Our analysis involved three stages:

First, an overall Pearson's Chi-squared test (without continuity correction) was performed on the full dataset (N = 2,306) to compare the total aggregate accuracy between DeepSeek-R1 and ChatGPT-4o.

Second, to compare the two models at a granular level (as presented in **Table 1** and **Table 2**), Fisher's exact test was applied. This test was used to analyze the 2x2 contingency table (Model [DeepSeek-R1 vs. ChatGPT-4o] vs. Outcome [Correct vs. Incorrect]) for each specific unit and year. Fisher's test was chosen for its accuracy in handling small sample sizes, which was particularly relevant for the MCQ analysis in Table 2 ($n \leq 10$).

Third, to analyze intra-model performance differences across knowledge domains (as discussed in response to reviewer 1.10), we conducted a 4-sample test for equality of proportions (prop.test) for each model. This test compared the accuracy of a single model across the four different units (Unit 1 vs. Unit 2 vs. Unit 3 vs. Unit 4) based on the aggregated 5-year data. Where this overall test was found to be significant, a pairwise comparison of proportions (pairwise.prop.test) with Bonferroni $p$-value correction was performed to identify which specific units differed significantly from one another.

**Qualitative Case Study Selection**

To supplement the quantitative accuracy data and provide deeper insight into the models' reasoning patterns and failure types, we added a qualitative analysis component. From the 2,306-item dataset, we employed purposive sampling to select two highly representative cases for in-depth analysis. The rationale for this selection was twofold: first, to provide a direct, qualitative explanation for our central quantitative finding—that the domain-specific model (DeepSeek-R1) significantly outperformed the general-purpose model (ChatGPT-4o) in both Unit 1 (Pharmaceutical Foundations) and Unit 4 (Clinical Integration); and second, to illustrate two distinct, high-stakes failure modes critical to pharmaceutical practice, in which the general-purpose model failed while the domain-specific model succeeded.

The first selected case (Case 1), drawn from Unit 1, involved a single-choice question requiring a direct comparison of antitussive potencies, allowing for the analysis of a factual knowledge deficit in foundational pharmacology. The second case (Case 2), a complex, case-based question from Unit 4, required the identification of an incorrect statement regarding Hormone Replacement Therapy (HRT)—a topic with historically conflicting guidelines. This case was selected to analyze a failure in complex clinical synthesis and guideline discernment. The objective results of these cases are presented in the Results section, and a full analysis of the models' reasoning is provided in the Discussion section.

**Results**

A Pearson's Chi-squared test was conducted to compare the overall accuracy of DeepSeek-R1 and ChatGPT-4o based on the full set of 2,306 prediction results. The test revealed a highly significant difference between the two models ($\chi^2(1, N = 2306) = 159.66, p < 2.2e-16$). DeepSeek-R1 achieved an overall accuracy of 90.0% (2076/2306), while ChatGPT-4o reached 76.1% (1754/2306). These results provide robust statistical evidence supporting the superior overall performance of DeepSeek-R1 on the Chinese Pharmacist Licensure Examination.

Table 1 reports the detailed unit-level accuracy across five years. The data shows that DeepSeek-R1's performance advantage was statistically significant in the vast majority of units, particularly in the fact-based Unit 1 (Pharmaceutical Foundations) and Unit 4 (Clinical Integration), as well as the regulation-focused Unit 2 (most $p < .05$). However, a notable exception was observed in Unit 3 (Dispensing and Prescription Review). In this unit, which assesses practical judgment, the performance difference between the two models was not statistically significant in four out of the five years (2017, 2018, 2020, 2021; all $p > .05$). This suggests that the models' capabilities are more evenly matched when tackling complex, judgment-based scenarios compared to factual recall.

Further analysis was conducted on the more challenging multiple-choice questions (MCQs), with results detailed in Table 2. While performance fluctuated between the models across different years and units (e.g., DeepSeek-R1 outperformed in 2017-Unit 3, while ChatGPT-4o outperformed in 2018-Unit 4), none of these unit-level differences reached statistical significance (all $p > .05$). This lack of statistical significance is likely attributable to the small sample size for MCQs within each unit ($n \leq 10$), which limits the statistical power to detect a true difference.

**Table 1** Overall Unit-Level Accuracy Comparison (2017–2021)

| Year | Unit | Deepseek-R1 Correct/Total (%) | ChatGPT-4o Correct/Total (%) | Total Items (n) | $p$-value |
|---|---|---|---|---|---|
| 2017 | Unit1 | 106/120 (88.33) | 60/120 (50.00) | 120 | <.001 |
|  | Unit2 | 113/120 (94.17) | 94/120 (78.33) | 120 | <.001 |
|  | Unit3 | 97/103 (94.17) | 89/103 (86.41) | 103 | 0.098 |
|  | Unit4 | 111/120 (92.50) | 84/120 (70.00) | 120 | <.001 |
| 2018 | Unit1 | 108/119 (89.92) | 68/119 (57.14) | 119 | <.001 |
|  | Unit2 | 109/119 (90.76) | 88/119 (73.95) | 119 | <.001 |
|  | Unit3 | 91/99 (91.92) | 87/99 (87.88) | 99 | 0.480 |
|  | Unit4 | 113/120 (94.17) | 90/120 (75.00) | 120 | <.001 |
| 2019 | Unit1 | 106/120 (88.24) | 78/120 (65.00) | 120 | <.001 |
|  | Unit2 | 111/120 (92.50) | 91/120 (75.83) | 120 | <.001 |
|  | Unit3 | 88/99 (88.89) | 74/99 (74.75) | 99 | 0.016 |
|  | Unit4 | 106/120 (88.33) | 90/120 (75.00) | 120 | 0.012 |
| 2020 | Unit1 | 93/120 (77.50) | 78/120 (65.00) | 120 | 0.045 |
|  | Unit2 | 105/120 (87.50) | 81/120 (67.50) | 120 | <.001 |
|  | Unit3 | 88/100 (88.00) | 79/100 (79.00) | 100 | 0.127 |
|  | Unit4 | 116/120 (96.67) | 89/120 (74.17) | 120 | <.001 |
| 2021 | Unit1 | 97/120 (77.50) | 74/120 (61.67) | 120 | 0.002 |
|  | Unit2 | 107/120 (90.76) | 82/120 (68.33) | 120 | <.001 |
|  | Unit3 | 95/107 (88.79) | 86/107 (80.37) | 107 | 0.129 |
|  | Unit4 | 116/120 (96.67) | 92/120 (76.67) | 120 | <.001 |

In addition to aggregate performance, we conducted year-wise statistical comparisons of multiple-choice question accuracy using Fisher's exact test, as detailed in Table 2. The results show that while DeepSeek-R1 consistently achieved higher multiple-choice accuracy than ChatGPT-4o in the earlier years (e.g., 2017-Unit 3, 2019-Unit 1), and ChatGPT-4o outperformed DeepSeek-R1 in later years (e.g., 2018-Unit 4, 2021-Unit 3 & 4), none of these unit-level differences reached statistical significance (all $p > 0.05$). This lack of significance is likely due to the small sample size for MCQs within each unit (n ≤ 10).

**Table 2** Unit-Level Accuracy Comparison on Multiple-Choice Questions (MCQs) (2017–2021)

| Year | Unit | Deepseek-R1 Correct/Total (%) | ChatGPT-4o Correct/Total (%) | Total MCQs (n) | $p$-value |
|---|---|---|---|---|---|
| 2017 | Unit1 | 7/10 (70.00) | 6/10 (60.00) | 10 | 1.000 |
|  | Unit2 | 8/10 (80.00) | 8/10 (80.00) | 10 | 1.000 |
|  | Unit3 | 6/8 (75.00) | 2/8 (25.00) | 8 | 0.132 |
|  | Unit4 | 6/10 (60.00) | 4/10 (40.00) | 10 | 0.656 |
| 2018 | Unit1 | 7/10 (70.00) | 6/10 (60.00) | 10 | 1.000 |
|  | Unit2 | 5/10 (50.00) | 3/10 (30.00) | 10 | 0.650 |
|  | Unit3 | 6/9 (66.67) | 5/9 (55.00) | 9 | 1.000 |
|  | Unit4 | 3/10 (30.00) | 7/10 (70.00) | 10 | 0.179 |

| Year | Unit | | | | |
|---|---|---|---|---|---|
| 2019 | Unit1 | 5/10 (50.00) | 1/10 (10.00) | 10 | 0.141 |
| | Unit2 | 5/10 (50.00) | 5/10 (50.00) | 10 | 1.000 |
| | Unit3 | 4/7 (57.14) | 3/7 (42.86) | 7 | 1.000 |
| | Unit4 | 3/10 (30.00) | 5/10 (50.00) | 10 | 0.650 |
| 2020 | Unit1 | 7/10 (70.00) | 5/10 (50.00) | 10 | 0.650 |
| | Unit2 | 5/10 (50.00) | 5/10 (50.00) | 10 | 1.000 |
| | Unit3 | 8/9 (90.00) | 5/9 (55.56) | 9 | 0.082 |
| | Unit4 | 8/10 (80.00) | 6/10 (60.00) | 10 | 0.628 |
| 2021 | Unit1 | 6/10 (60.00) | 3/10 (30.00) | 10 | 0.370 |
| | Unit2 | 7/10 (70.00) | 8/10 (80.00) | 10 | 1.000 |
| | Unit3 | 3/9 (33.33) | 6/9 (66.67) | 9 | 0.347 |
| | Unit4 | 9/10 (90.00) | 10/10 (100.00) | 10 | 1.000 |

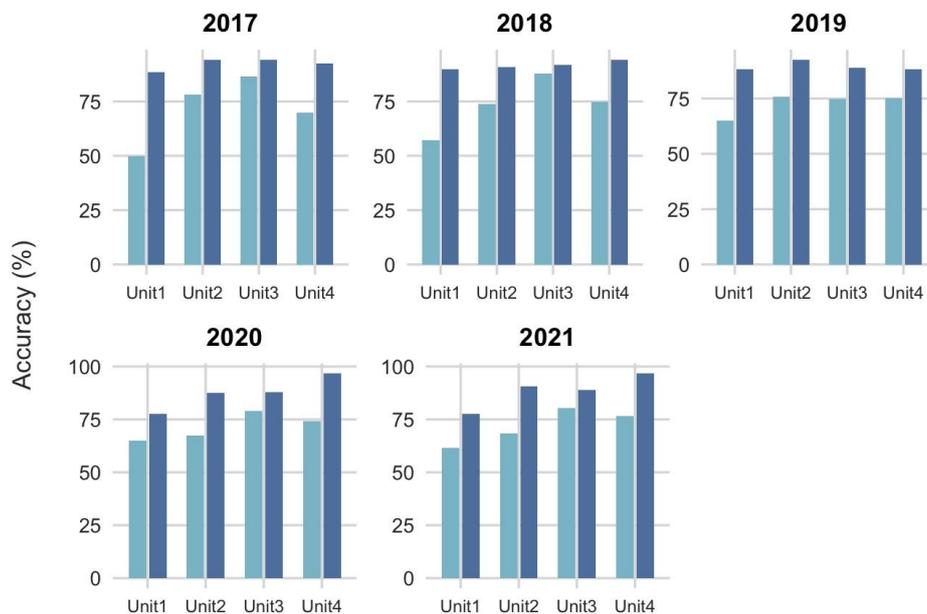

**Figure 1. Year-wise Unit-Level Overall Accuracy Comparison (2017–2021)**

**Figure 1.** shows the overall unit-level accuracy (%) of DeepSeek-R1 and ChatGPT-4o, corresponding to the data in **Table 1**. Each panel represents a different year. DeepSeek-R1 consistently outperforms ChatGPT-4o in most units, particularly in Unit 1 (pharmacy foundation) and Unit 4 (clinical synthesis), where the differences were found to be statistically significant (see Table 1).

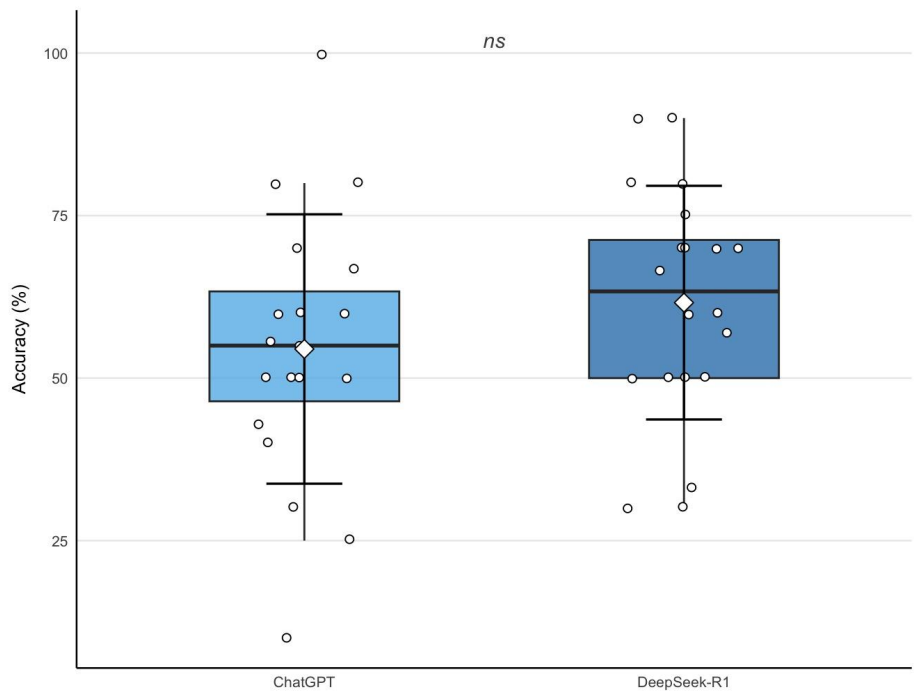

**Figure 2.** Distribution of Unit-Level Accuracy in Multiple-Choice Tasks (2017–2021)

**Figure 2.** represents the distribution of unit-level accuracy (%) for DeepSeek-R1 and ChatGPT-4o on Multiple-Choice Questions (MCQs), corresponding to the data in **Table 2**. Each dot denotes one unit-year sample. While the median accuracy for DeepSeek-R1 appears higher, the overall difference between the two models on MCQ tasks was not statistically significant ($p > 0.05$ for all unit-level comparisons, as shown in Table 2), which is denoted by "ns" in the plot.

**Figure 3.** Case 1: Pharmaceutical Foundations (Antitussive Potency)

In **Figure 3.**, a comparison of model responses to a question on the pharmacological potency of antitussives. The correct answer is D (Benproperine > Codeine > Pentoxyverine). The figure shows ChatGPT-4o's incorrect response (A) and its reasoning, contrasted with DeepSeek-R1's correct response (D) and its detailed pharmacological analysis.

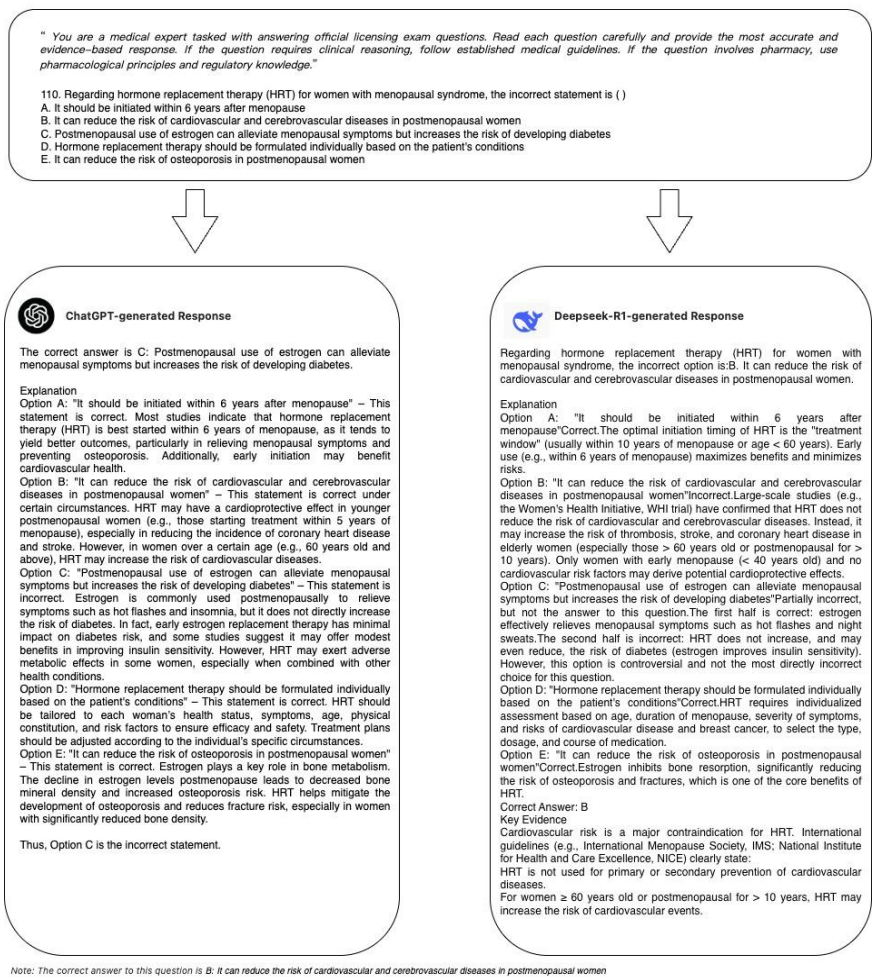

**Figure 4.** Case 2: Clinical Integration (HRT Guideline Discernment)

As shown in **Figure 4.**, a comparison of model responses to a case-based question requiring the identification of an incorrect statement about Hormone Replacement Therapy (HRT). The correct answer is B (the statement that HRT reduces cardiovascular risk). The figure shows ChatGPT-4o's incorrect response (C), contrasted with DeepSeek-R1's correct response (B). As shown in its explanation, DeepSeek-R1's reasoning explicitly cited the Women's Health Initiative (WHI) trial.

**Discussion**

This study explores the performance differences between large language models (LLMs) on the Chinese Pharmacist Licensing Examination, aiming to answer a fundamental question: can these technologies genuinely serve pharmaceutical education, rather than merely helping users pass exams? By examining different exam units, we gain a real-world educational perspective on where LLMs excel and where they fall short.

In Unit 1 (Pharmaceutical Foundations), DeepSeek-R1 showed steady and significant advantages over ChatGPT-4o. We attribute this performance gap primarily to the fundamental differences in their training architecture and data corpora. DeepSeek-R1 is a model specifically optimized with domain-specific (Chinese medical and pharmaceutical) materials [16]. Its architecture is thus highly aligned with the exam's specialized terminology, factual recall, and contextual nuances. In contrast, ChatGPT-4o is a general-purpose model. While powerful, its knowledge base is vastly broader and less concentrated, with a documented English-centric bias. This may limit its precision when handling the highly specific, non-English terminology found in the NPLE. The structured nature of these foundational questions (e.g., pharmacology, chemistry), which rely on clear, factual answers, directly benefits DeepSeek-R1's specialized training. The exam questions in this section tend to follow a clear structure, and the answers are usually straightforward. This format naturally favors systems that are good at handling factual content. In Chinese pharmacy education, there's long been concern that too much emphasis is placed on memorization, while deeper understanding often takes a back seat. Given this, certain digital tools might be useful for organizing key ideas or helping students review before exams.

Interestingly, despite DeepSeek-R1's clear advantage over ChatGPT-4o in this unit, our intra-model analysis revealed that Unit 1 (85.1%) was statistically DeepSeek-R1's weakest-performing category, scoring significantly lower than Unit 2 ($p = .015$) and Unit 4 ($p < .001$). This trend was even more pronounced for ChatGPT-4o, for which Unit 1 (59.8%) was by far its worst-performing category ($p < .001$ compared to all other units). This suggests that the highly specific, foundational knowledge required by Unit 1 may be a key bottleneck for both domain-specific and general-purpose LLMs.

In Unit 2, which covers pharmaceutical legislation and ethics, the performance gap between systems became less noticeable. That's not surprising—this part of the curriculum isn't about fixed answers, but about interpreting legal language and making sense of region-specific regulations, which are constantly evolving. In the classroom, teachers already face this challenge: policies change from one province to another, and what's correct in one context may not apply elsewhere. So it makes little sense to expect one perfect answer in such cases. A better use of technology here might be to highlight important points or lay out different ways of interpreting a rule, while the job of explaining the real-world context still falls to the teacher.

In Unit 3, which deals with dispensing and prescription review, the results were particularly revealing and challenge the notion that AI struggles with "judgment". Contrary to expectations, this unit was ChatGPT-4o's statistically strongest category (81.7%), performing significantly better than both Unit 2 ($p = .004$) and Unit 4 ($p = .021$). For DeepSeek-R1, its performance (90.4%) was also robust, statistically indistinguishable from its other top-performing units (Unit 2 and 4). This suggests that the "judgment" tested by the NPLE in this unit may be less about abstract ethical reasoning and more about the application of complex, rule-based patterns (e.g., identifying known contraindications or dosing errors), a task at which ChatGPT-4o excels.

Unit 4, focused on clinical integration, confirmed our hypothesis for DeepSeek-R1 but not for ChatGPT-4o. For DeepSeek-R1, Unit 4 was its highest-scoring category (93.7%), performing significantly better than its foundational Unit 1 ($p < .001$). This aligns perfectly with a model specifically trained for advanced, domain-specific clinical synthesis. For ChatGPT-4o, however, performance in Unit 4 (74.2%) was statistically weaker than its top-performing Unit 3 ($p = .021$) and statistically indistinguishable from its performance in Unit 2 ($p = 1.000$). This indicates that while DeepSeek-R1's strength lies in its specialized clinical integration, ChatGPT-4o's advantage lies more in high-level, rule-based case analysis (Unit 3).

Evident when comparing the high overall accuracy (**Table 1**) with the specific MCQ performance (**Table 2**), is the significant performance gap between single-choice items and the more complex multiple-choice questions (MCQs). Both models achieved substantially lower accuracy on MCQs (DeepSeek-R1: 58.9%; ChatGPT-4o: 53.6%) compared to their performance on the predominantly single-choice items (DeepSeek-R1: 90.0%; ChatGPT-4o: 76.1% ).

This performance drop highlights that while LLMs are proficient at fact retrieval (which dominates single-choice questions), they still face significant hurdles in complex, multi-step synthesis and evaluation (the core requirement of MCQs).

Furthermore, looking within the MCQ tasks (**Table 2**), our statistical analysis revealed no significant difference between the two models in any specific unit (all $p > .05$). This lack of significance, as noted in the Results , is likely due to the small sample size ($n \leq 10$) in each sub-category, which makes it difficult to detect statistically valid trends. This highlights the sensitivity of these tools to question style or wording, but also underscores the need for larger, specialized MCQ datasets to draw firm conclusions about performance on these more complex items.

Beyond the quantitative scores, the qualitative case studies (**Figures 3** and **4**) provide crucial insights into the models' reasoning and failure modes. In Case 1 (**Figure 3**), ChatGPT-4o's failure to rank antitussive potency suggests a clear deficit in its foundational pharmacological knowledge base. Its explanation, while fluent, confounds the drugs' properties. DeepSeek-R1's correct response demonstrates a precise factual alignment consistent with its domain-specific training. Case 2 (**Figure 4**) is even more revealing. The HRT question tests the ability to discern modern, high-risk clinical guidelines from historically conflicting information. ChatGPT-4o's incorrect answer (C) suggests its broad, general-data training may have exposed it to obsolete "facts" (e.g., pre-WHI trial beliefs) or that it failed to prioritize the most critical, high-risk contraindication (cardiovascular risk). In contrast, DeepSeek-R1's correct answer (B) and its explicit reference to the WHI trial demonstrate a reasoning process aligned with current, expert-level clinical consensus.

The potential for model failure, as starkly illustrated by the HRT case (Figure 4), raises significant ethical considerations regarding the deployment of LLMs in high-stakes training and credentialing systems. Our findings challenge the notion that high quantitative scores (like ChatGPT-4o's 76.1%) equate to trustworthiness. A model that cannot differentiate current, high-risk clinical guidelines from outdated information poses a direct pedagogical and clinical risk. This highlights the danger of model misuse, where educators or learners might over-rely on a flawed tool, and points to the critical need for accountability frameworks. Furthermore, the "black-box" nature of these models remains a barrier to explainability. Our qualitative analysis (**Figures 3 & 4**) serves as a limited form of probing, but it also reinforces that, ethically, LLMs should not act as infallible judges. They are better positioned as "suggesters" or "reflectors" that propose options with reasoning trails, prompting learners to engage in critical reflection.

More broadly, this raises a practical question for educators working with smart tools: can we match different types of model support to different types of content? For instance, Units 1 and 4, with their emphasis on facts and clinical reasoning, might be well suited for automated question generation or grading. In contrast, Units 2 and 3, where legal interpretation and professional judgment come into play, still call for human input—both to guide learning and to avoid oversimplifying what are often complex decisions.

In the Chinese context, real-world deployment also demands alignment with continuing education and remote training platforms. Many county-level pharmacists must complete annual online certification. These systems often suffer from inconsistent evaluation standards, overworked instructors, and slow feedback cycles. A model like DeepSeek-R1, with high controllability, could generate personalized quizzes, stage-based feedback, or peer benchmarking—helping platforms become fairer and more scalable.

However, these models still require human supervision. In Units 2 and 3, ambiguity in policy interpretation, regional variation, and case-based complexity can easily mislead the model. Ethically, LLMs should not act as judges but as suggesters or reflectors. For example, instead of outputting definitive answers, the model might propose several plausible options with reasoning trails, prompting learners to reflect. This also addresses key concerns regarding AI 'interpretability' and 'fairness'; rather than presenting an opaque, definitive 'correct' answer, this 'suggester' role provides a more transparent and equitable educational tool that empowers human judgment instead of replacing it.

Also, response efficiency must not become the sole measure of educational success. We are concerned that if models are widely used in feedback systems, students may gradually lose interest in understanding the context of problems and instead become fixated on identifying the correct options. This would deviate from the core goals of medical and pharmaceutical education. Therefore, we suggest enhancing model outputs with modules such as counterexample analysis, scenario-based comparisons, and alerts on regulatory ambiguities. These features would guide learners to develop integrative judgment and a sense of professional responsibility, rather than simply improving test scores.

At the time of testing (February–March 2025), DeepSeek-R1—released on January 20, 2025—represented the latest publicly available model with task-specific optimization. In contrast, OpenAI had not yet released a chain-of-thought-enhanced, Chinese-adapted ChatGPT-4o by that time. This version asymmetry reflects real-world deployment constraints and captures a meaningful cross-section of then-available tools.

**Conclusion**

DeepSeek-R1 demonstrated robust alignment with the structural and semantic demands of the Chinese Pharmacist Licensing Examination. These findings suggest that domain-specific models warrant further investigation as potential tools in pharmacy-related educational settings. While this highlights the potential for targeted LLMs in digital health education, it also reinforces the absolute necessity of human oversight, particularly in legally and ethically sensitive contexts.

However, these findings must be interpreted with caution. The superiority of the domain-specific model (DeepSeek-R1) highlights the challenge of generalizability, and its performance on this specific, text-only Chinese exam may not translate to other professional contexts, languages, or real-world multimedia clinical scenarios. This balanced perspective—acknowledging both the potential and the limitations—is essential for the responsible integration of AI into public health training.

**Limitations**

Several limitations should be acknowledged. First, and most importantly, this study lacks a direct human performance benchmark. Granular data on human performance (e.g., average scores per unit, per year) for the Chinese NPLE is not publicly available. Therefore, our findings do not assess whether these models "pass" the exam relative to a human cohort, but rather compare their performance against each other and across different knowledge domains. Future research should aim to acquire such data, if possible, to provide a more definitive benchmark.

Second, our methodology, by design, evaluated the models' default 'real-world' performance, which has a known limitation in reproducibility. The experiment was conducted using the publicly available web interfaces, and as such, specific underlying parameters were deliberately not altered from the default, as the goal was to evaluate the models' standard performance as encountered by a typical user. Default settings on these platforms typically involve a non-zero temperature, which introduces stochasticity. Therefore, our results represent a snapshot of performance rather than a fully deterministic benchmark. However, we believe the substantial 14-percentage-point gap in overall accuracy (90.0% vs. 76.1%) is highly unlikely to be an artifact of this randomness and reflects a true difference in the models' underlying capabilities.

Third, the study focuses solely on the Chinese Pharmacist Licensure Examination, which may limit the generalizability of the findings to other medical or pharmaceutical assessments. Fourth, our study used questions in their original Chinese format. This may have inherently favored DeepSeek-R1, a model optimized for Chinese, while potentially disadvantaging ChatGPT-4o due to its known English-centric bias and potential internal translation errors. Fifth, the analysis does not account for the potential impact of question difficulty variations across years.Lastly, our qualitative analysis of the underlying reasoning processes was exploratory and limited to two representative cases . We did not perform a systematic, large-scale qualitative error analysis across all 2,306 items. Such a comprehensive review remains a critical direction for future work.

**References**
[1] Fang Y, Yang S, Zhou S, Jiang M, Liu J. Community pharmacy practice in China: past, present and future. *Int J Clin Pharm*. 2013;35:520-8.
[2] Yao D, Xi X, Huang Y, Hu H, Hu Y, Wang Y, et al. A national survey of clinical pharmacy services in county hospitals in China. *PLoS One*. 2017;12(11):e0188354.
[3] Qin SB, Zhang XY, Fu Y, Nie XY, Liu J, Shi LW, et al. The impact of the clinical pharmacist-led interventions in China: A systematic review and Meta-Analysis. *Int J Clin Pharm*. 2020;42:366-77.
[4] Bian Y, Yang ZY, Xiong Y, Tong RS, Yan JF, Long EW. Discussion on clinical pharmaceutical service model in prevention and treatment of corona virus disease 2019. *Chin J New Drugs Clin Rem*. 2020;39(4):212-7.
[5] Gao Y, Xu T, Jin ZH, Wang SM, Long GL, Zheng ML. Practice and discussion of outpatient pharmaceutical care based on medical network model during 2019 novel coronavirus disease (COVID-19). *Chin J Hosp Pharm*. 2020;40:606-11.
[6] Meng L, Huang J, Qiu F, Sun S. Roles of the Chinese clinical pharmacist during the COVID‐19 pandemic. *J Am Coll Clin Pharm*. 2020;3(5):866.
[7] Wen R, Xu L, Zhu W. Suggestions on the revision of pharmacist examination outline for national professional qualification examination of licensed pharmacists. *药学实践与服务 (Pharmacy Practice and Service)*. 2022;40(4):383-6.
[8] Chang Y, Wang X, Wang J, Wu Y, Yang L, Zhu K, et al. A survey on evaluation of large language models. *ACM Trans Intell Syst Technol*. 2024;15(3):1-45.

[9] Raiaan MAK, Mukta MSH, Fatema K, Fahad NM, Sakib S, Mim MMJ, et al. A review on large language models: Architectures, applications, taxonomies, open issues and challenges. *IEEE Access*. 2024;12:26839-74.

[10] Mikac M, Horvatić M, Logožar R, Dumić E. ChatGPT-4o in Education-Use Cases in an Introductory Web Programming Course. *In: INTED2024 Proceedings*; 2024. p. 3173-82.

[11] Dave T, Athaluri SA, Singh S. ChatGPT-4o in medicine: an overview of its applications, advantages, limitations, future prospects, and ethical considerations. *Front Artif Intell*. 2023;6:1169595.

[12] Sallam M. ChatGPT-4o utility in healthcare education, research, and practice: systematic review on the promising perspectives and valid concerns. *Healthcare (Basel)*. 2023;11(6):887.

[13] Singhal K, Tu T, Gottweis J, Sayres R, Wulczyn E, Amin M, et al. Toward expert-level medical question answering with large language models. *Nat Med*. 2025;1-8.

[14] Huang J, Chang KCC. Towards reasoning in large language models: A survey. *arXiv*. 2022. arXiv:2212.10403.

[15] Longwell JB, Hirsch I, Binder F, Conchas GAG, Mau D, Jang R, et al. Performance of large language models on medical oncology examination questions. *JAMA Netw Open*. 2024;7(6):e2417641.

[16] Qin C, Zhang A, Zhang Z, Chen J, Yasunaga M, Yang D. Is ChatGPT-4o a general-purpose natural language processing task solver?. *arXiv*. 2023. arXiv:2302.06476.

[17] Bi X, Chen D, Chen G, Chen S, Dai D, Deng C, et al. Deepseek llm: Scaling open-source language models with longtermism. *arXiv*. 2024. arXiv:2401.02954.

[18] Fergus S, Botha M, Ostovar M. Evaluating academic answers generated using ChatGPT-4o. *J Chem Educ*. 2023;100(4):1672-5.

[19] Kung TH, Cheatham M, Medenilla A, Sillos C, De Leon L, Elepaño C, et al. Performance of ChatGPT-4o on USMLE: potential for AI-assisted medical education using large language models. *PLoS Digit Health*. 2023;2(2):e0000198.

[20] Chalkidis I. ChatGPT-4o may pass the bar exam soon, but has a long way to go for the lexglue benchmark. *arXiv*. 2023. arXiv:2304.12202.

[21] Li KC, Bu ZJ, Shahjalal M, He BX, Zhuang ZF, Li C, et al. Performance of ChatGPT-4o on Chinese master's degree entrance examination in Clinical Medicine. *PLoS One*. 2024;19(4):e0301702.

[22] Yildirim-Erbasli SN, Bulut O. Conversation-based assessment: A novel approach to boosting test-taking effort in digital formative assessment. *Comput Educ Artif Intell*. 2023;4:100135.

**Declaration**

**Funding:** This research is funded by the Shanghai Municipal Health Commission (Grant No. 2024WF03).

**Ethics, Consent to Participate, and Consent to Publish:** Not applicable. This study did not involve human subjects or personal data. The exam questions were obtained from publicly accessible repositories and have been previously utilized in peer-reviewed research.

**Data Availability:** The datasets generated and analysed during the current study are available in the supplementary file submitted with the manuscript.

**Author Contributions:** Xinran WANG is the first author and led the entire study. She was responsible for the study design, data analysis, and manuscript drafting. Boran ZHU contributed to model implementation, result validation, and manuscript revision. Shujuan ZHOU participated in data cleaning and literature review. Ziwen LONG provided support in figure preparation and methodological consultation. Shu ZHANG is the lead corresponding author and supervised the research process, provided critical insights, and guided manuscript development. Dehua ZHOU is the second corresponding author and contributed to study design and final revision of the manuscript.

**Guarantor:** Shu ZHANG is the guarantor of this article.

**Acknowledgement:** This work was supported by the Shanghai Municipal Health Commission, whose support is gratefully acknowledged.